# HYBRID APPROACHES FOR AUTOMATIC VOWELIZATION OF ARABIC TEXTS

Mohamed Bebah[1] Chennoufi Amine[2] Mazroui Azzeddine[3] and Lakhouaja Abdelhak[4]

[1]Arab Center for Research and Policy Studies, Doha, Qatar
[2]Faculty of Sciences/University Mohamed I, Oujda, Morocco
[3]Faculty of Sciences/University Mohamed I, Oujda, Morocco
[4]Faculty of Sciences/University Mohamed I, Oujda, Morocco

***ABSTRACT***

*Hybrid approaches for automatic vowelization of Arabic texts are presented in this article. The process is made up of two modules. In the first one, a morphological analysis of the text words is performed using the open source morphological Analyzer AlKhalil Morpho Sys. Outputs for each word analyzed out of context, are its different possible vowelizations. The integration of this Analyzer in our vowelization system required the addition of a lexical database containing the most frequent words in Arabic language. Using a statistical approach based on two hidden Markov models (HMM), the second module aims to eliminate the ambiguities. Indeed, for the first HMM, the unvowelized Arabic words are the observed states and the vowelized words are the hidden states. The observed states of the second HMM are identical to those of the first, but the hidden states are the lists of possible diacritics of the word without its Arabic letters. Our system uses Viterbi algorithm to select the optimal path among the solutions proposed by Al Khalil Morpho Sys. Our approach opens an important way to improve the performance of automatic vowelization of Arabic texts for other uses in automatic natural language processing.*

***KEYWORDS***

*Arabic language, Automatic vowelization, morphological analysis, hidden Markov model, corpus*

## 1. INTRODUCTION

The Arabic writing system is characterized in most texts by the lack of diacritics: i.e., short vowels /a/ "ـَ" (fatha), /u/ "ـُ" (damma) and /i/ "ـِ" (kasra), in addition to signs /an/"ـً"(tanween al-fatha), /un/ "ـٌ" (tanween al-damma ), /un/ "ـٍ"(tanween al-kasra), consonant doubling "ـّ" (shadda) and vowel absence "ـْ" (sukuun). The absence of these signs generates a significant increase of the ambiguity in the Arabic text, which can cause confusion in more than 90% of the text words [1]. Despite the fact that the reader with a certain level of knowledge of Arabic language can easily recover the missing diacritics by using the context of words and its knowledge of the morphology and the syntax of the Arabic language, texts without diacritics present an obstacle for non-native learners of the Arabic language and those with learning difficulties. Similarly, the limits of performance of several applications of natural language processing for Arabic (NLPA) such as parsers and Treebank are in part a result of the absence of diacritics in Arabic texts [2,3]. Indeed, unlike European languages where it is easy to identify oral phonemes corresponding to texts (Text to Speech), it is imperative for Arabic texts to retrieve the diacritics before researching the correspondent oral phonemes [4]. On the other hand, some research has underlined the importance of using texts with diacritics to increase the efficiency of speech recognition [5].

Given the importance of the recovery of diacritics, several attempts have been made by research teams over the past two decades. They developed automatic vowelization algorithms for Arabic texts with an acceptable efficiency. These attempts can be divided, as most applications of NLP, in two categories: the first one concerns commercial companies that have developed automatic vowelization algorithms as an independent program or as a part of another application, such as the text to speech or spell checker. Among the most interesting projects, there is the vowelization program called ArabDiac developed by Egyptian factory RDI (Research & Development International) and those developed by SAKHR Software (Egyptian company)) and CIMOS company. The company Google has also launched three years ago Tashkeel program. It was a free automatic vowelization program for Arabic texts, but this service was discontinued for unknown reasons. Despite the importance of these attempts, their commercial and monopolistic nature, which prevents access to the source code and linguistic resources used in the development of these programs, does not allow to improve or to integrate them into other applications. The second category of these attempts is due to the efforts of researchers in projects within academic research centers. These efforts have resulted in the past two decades the emergence of many attempts in this field. The used approaches are often the statistical type based on the Markov models or the n-gram models [6,7,8,9]. However, some hybrid approaches using linguistic analysis followed by a statistical treatment are also used to develop an automatic vowelization system [4,10,11,12].

An automatic vowelization systems based on a hybrid approaches which combine a morphological analysis and hidden Markov models is presented in this article. These approaches differ from other hybrid approaches to linguistic and statistical levels. Indeed, at the linguistic level, the open source morphological Analyzer Alkhalil Morpho Sys is used (*www.sourceforge.net/projects/alkhalil*) [13]. The integration of this Analyzer in our vowelization system required the addition of a lexical database containing the most frequent words in order to adjust their diacritics and circumvent the problems of slow due to the high number of solutions proposed by the morphological Analyzer. This database has been generated from the corpora of more than 250 million words of eight Arabic corpuses available on the Internet. Our system uses the results of the morphological Analyzer and extracts possible vowelizations out of context proposed for words. Statistically, two Markov models are used. The first model uses the unvowelized Arabic words as observed states and vowelized words as hidden states, while the second model keeps the same observed states of the first Markov model, but the hidden states are the lists of possible diacritics of word without their Arabic letters. Finally, a representative corpus is used in the training phase.

The article is organized as follows. In Section 2, the state of art in the field of Arabic vowelization is recalled. After, the morphological Analyzer Alkhalil Morpho Sys used in the first part of our system is presented in Section 3, and the statistical approach adopted in the second part is explained. The fourth and fifth sections are devoted respectively to training and testing phases. Finally, conclusion and future prospects are given in the last section.

## 2. State of the Art

Referring to the previous works, the approaches related to the automatic vowelization of Arabic texts can be divided into three sections: rule-based approaches, statistical approaches and hybrid approaches.

## 2.1. Rule-based Approaches

There have been some researches aimed at programming audio, morphological, grammatical and spelling grammar rules in order to vowelize the Arabic words. Among the first studies on the subject, there is the approach mentioned in [14] that uses syntactic rules for semi-automatic vowelization of Arabic verbs. In [1], Debili, Fathi and Hadhemi Achour present a study of the automatic vowelization Arabic texts in relation to the ambiguity of written words and the impact of lexical and morphological analysis and POS tagging in detecting ambiguities and hence the vowelization of words of Arabic texts.

But given the high level of ambiguity, the large number of morphological and syntactic rules, the unavailability of an efficient parser and the need of a semantic analysis in some cases, it is difficult to develop an automatic vowelization system based solely on grammar rules and this explains the absence of such an efficient system.

## 2.2. Statistical Approaches

Given the great feat achieved by the statistical approaches in various fields of natural language processing such as automatic speech recognition, machine translation and information retrieval [15], the majority of works in the field of automatic vowelization of Arabic texts have adopted these approaches. The used methods are quite varied. Indeed, some researchers have developed statistical methods based on research of diacritic marks on the character level. Others have exploited these methods to identify the diacritic marks on the word level. Finally, a third group of researchers have developed hybrid methods coupling the two approaches.

Indeed, Emam and Fischer [7] had filed a patent for an automatic vowelization system for Arabic texts based on the example-based machine translation. In order to select the most probable sentence vowelization, the approach consists first of searching in a database some examples of the sentence, then the parts of the sentence and finally the isolated words of the sentence. In addition, they apply statistical methods (n-gram model) at characters level for words that do not appear in the database. Not far from this idea, there is in [8] an approach based on the statistical machine translation technology based on parallel corpus composed of vowelized texts and their unvowelized counterparts. Also, Gal [9] has presented a Markovian approach to vowelize the Arabic and Hebrew texts. The author has used in this work the Holy Quran texts in the Arabic language and the Bible texts in the Hebrew language. In this work, the unvowelized words and the vowelized words represent, respectively the observation and the hidden states of the model. However, the Arabic vowelization in this work is limited to short vowels.

For automatic vowelization by the Markovian approaches, there are works of Deltour [16] and Elshafei, Al-Muhtaseb, and Alghamdi [6]. In [16], the researcher reviewed some statistical methods of automatic vowelization on the level of characters and words, and concluded that the best results are obtained when she has used the hidden Markov models. Similarly, the researchers have presented in [6] a Markovian approach distinct from those of work of Gal [9] since it concerns all diacritical marks, and the training and evaluation processes were realized by using a variety of texts and not only the Quranic texts.

In [17], the authors have presented an Arabic diacritizer developed at King Abdel Aziz City for Science and Technology (KACST). This system is based on a quad-grams applied at the character level.

### 2.3. Hybrid Approaches

Hybrid methods are algorithms that combine linguistic rules and statistical processes to exploit the strengths of both approaches. From important work in this field, there is the diacritizer ArabDiac developed by RDI [18]. This system uses the morphological Analyzer ArabMorpho and the POS tagger ArabTagger in a probabilistic framework where the choice of the best vowelization is performed using the A* search algorithm.

The authors Nelken and Shieber [10] have proposed a method for automatic vowelization based on finite state automata. This method combines statistical tri-gram models at the word level, quad-gram models at the character level and a slight morphological Analyzer that recognizes the prefixes and suffixes of words.

Zitouni, Sorensen, and Sarikaya [3] presented a vowelizer system using statistical classifier based on maximum entropy. Morpho-syntactic information is used to select the best classification.

In [4], diacritics of Arabic words are restored by combining a morphological analysis and contextual information with an acoustic model. Automatic vowelization in this work is seen as an unsupervised tagging problem where each word is tagged as one of the possible solutions provided by the Buckwalter Analyzer [19]. The maximum expectation algorithm was used in this work to perform the learning section.

In [11], there is a similar approach to that presented by Vergyri and Kirchhoff [4] in the sense that the vowelization problem was seen as a problem of choosing the best solution among those proposed by the Buckwalter Analyzer.

We have already presented in [12] a vowelization system for Arabic sentences based on a morpho-statistical approach. The Analyzer Alkhalil Morho Sys was used to identify the various possible vowelized patterns of words analyzed out of context. Then, the statistical treatment removes ambiguity by considering these patterns as hidden states of a hidden Markov model. It remains to note that this approach shares with the approach presented in this research the utilization of the Analyzer Alkhalil Morpho Sys in the first step. However, the Alkhalil version used in this research was modified by adding a lexicon of the most frequent Arabic words. In addition, the hidden states for each approach and the used corpus in the learning and testing phases are different.

## 3. PRESENTATION OF THE SYSTEM

This section give a detailed presentation of the developed system. The process of automatic vowelization of Arabic texts will be done in two main phases, as shown in Figure 1.

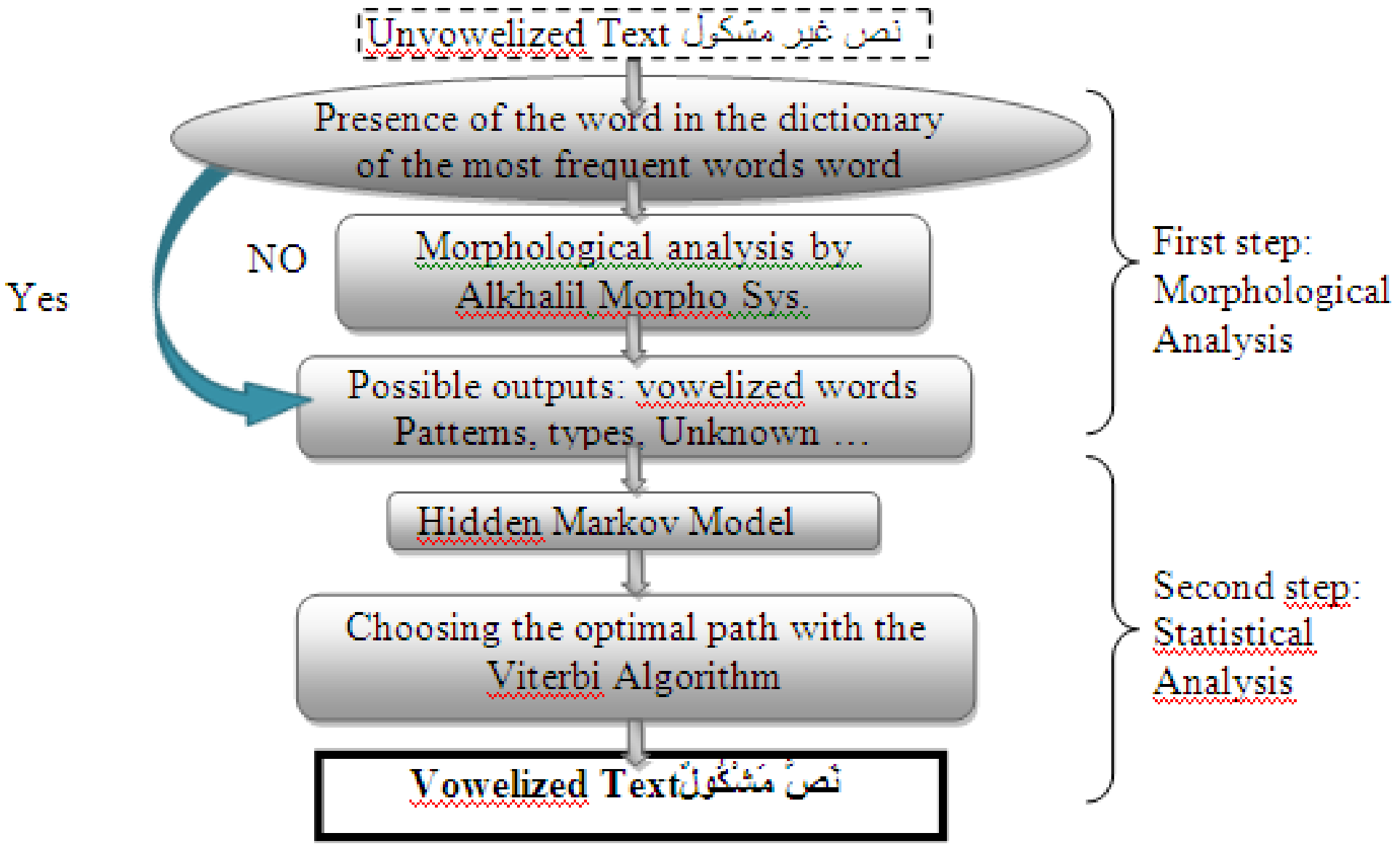

Figure 1. Steps for automatic vowelization of Arabic texts

## 3.1. First Phase: Morphological Analysis

Morphological analysis is performed using the open source Analyzer Alkhalil Morpho Sys. It provides all possible vowelizations for each word of the text taken out of context. To further clarify this point, a brief overview of the Analyzer and its mode of operation are given at the beginning, and then the detail of the lexical database that is integrated in the Analyzer databases is presented. To evaluate the contribution of the added lexical database, the performances of the system with and without the lexicon are tested and the results will be presented in the section related to tests bellow.

### 3.1.1. Overview of Alkhalil Morhp Sys

The morphological Analyzer Alkhalil Morpho Sys is considered one of the most important open source Analyzers [13]. It was conducted as part of a partnership project between the Mohammed First University Oujda Morocco (UMP), the Arab League Educational, Cultural and Scientific Organization (ALECSO) and the King Abdulaziz City for Science and Technology (KACST). For a given word, Alkhalil Morpho Sys identifies all possible solutions associated with their morpho-syntactic features as defined below:

- Possible vowelizations of the word.
- Possible proclitics and enclitics.
- Nature of the word (noun or verb or tool word).
- Vowelized patterns (for derivable words).
- Stems.
- Roots (for derivable words).
- Part of speech (for nouns and verbs).

The main steps of the morphological analysis for each word with Alkhalil Morpho Sys can be summarized as follows:

1. First step: it aims to identify potential prefixes and suffixes of the word by making all possible segmentations and checking if the first and the last segments of the segmentation belong respectively to databases of prefixes and suffixes. Due to the inflectional nature of the Arabic, this process often leads to more than one segmentation.
2. Second step: after removing the prefix and suffix associated with each valid segmentation of the first step, the remaining segment (stem) will be analyzed as a proper noun. To do this, the program uses a list of 6,000 proper nouns.
3. Third step: the program checks after if the stem belongs to the tool word list.
4. Fourth step: the program checks if the stem is a derived noun. This check will be done according to the following steps:

   - Identify possible patterns of the stem and compare them with those of the nominal patterns database.
   - Extract the possible roots of the stem and ascertain their presence in the roots database of the Arabic language.
   - Check the compatibility of each root with the pattern by referring to the database of roots accompanied by patterns of names derived from this root.

5. Step Five: The program checks if the stem is a verb. This check will be using the same steps as for the names.

These are briefly the most important steps of morphological analysis in the official version of Alkhalil Morpho Sys available on the website sourceforge. However, the integration of the Analyzer Alkhalil Morpho Sys in the automatic vowelization system obliged us to make adjustments that consist primarily of adding a database in the form of a dictionary of the Arabic words most commonly in the available Arab corpuses. Similarly, the mechanism of the morphological analysis outputs are modified, keeping only the possible vowelizations of words and ignoring the other information such as pattern, root, etc... In what follows, the detail of this dictionary is presented.

#### 3.1.2. Dictionary of the Most Frequent Words

This dictionary has been built on the one hand to accelerate the process of morphological analysis and on the other hand to adjust the quality of vowelization of the most frequent words belonging to the various texts. To do this, the steps bellows are followed:

- Gather a large Arabic corpus of more than 250 million words drawn from eight Arab corpuses available on the Internet. The corpuses are as follows:

  - The corpus of the researcher Ahmed Abdelali (*http://aracorpus.e3rab.com/ argistestsrv.nmsu.edu/AraCorpus/Data/*) which contains over 147 million unvowelized words. This corpus was constituted from a variety of media texts collected from 28 journalistic websites covering most Arab countries.
  - The corpus (*Tashkeela*) (*http://sourceforge.net/projects/tashkeela/*) which contains over 60 million of vowelized words and is composed of classical Arabic texts collected by the researcher T. Zerouki from library books (*Achamila*).
  - The open source Arabic corpora (OSAC) (*https://sites.google.com/site/motazsite/Home/ osac*) collected by researchers W. Moataz K. Saad and Ouissam Ashour in order to use it in applications of text mining. This corpus contains about 20 million unvowelized words gathered from local and international Arabic news websites.

- The corpus of L. Sulaiti (*http://www.comp.leeds.ac.uk/eric/latifa/research.htm*) composed of unvowelized contemporary texts covering several areas (political, economic, religious, sports, etc.). This corpus contains over half a million words.
- The corpus (*Khaleej-2004*) (*http://sourceforge.net/projects/arabiccorpus/files/*) collected by researchers M. Abbas and K. Smaili from newspaper (*Akhbar Al Khaleej*), Bahrain. This corpus, which contains about 2.8 million words, has been used in applications of automatic classification of documents [20].
- The UN Parallel Corpora (*http://www.uncorpora.org/*) which consists of 2,100 resolutions adopted by the General Assembly of the United Nations and written in the six official languages of the United Nations. Only the Arabic content of this resolutions containing about 2.5 million words has been processed. This corpus was originally collected for use in applications such as machine translation [21].
- The corpus of the RDI (*http://www.rdi-eg.com/RDI/TrainingData/*) society is composed of vowelized texts gathered mainly from classical Arabic books and a small percentage of contemporary writing. This corpus was collected in order to use in the field of automatic vowelization. It contains 20 million words.
- The ArabicWritten Corpus NEMLAR (Network for Euro-Mediterranean LAnguage Resources) (*http://catalog.elra.info/product_info.php?products_id=873*) was carried out by several researchers in the NEMLAR project [22]. It is a vowelized corpus, with morpho-syntactically tagging and marketed by European Language Resources Association (ELRA). It contains texts from various fields and consists of approximately half a million words.

- After the collection phase, the encoding and storing methods of different files are standardized using CP1256 code and ".txt" extension for backup files. These files are cleaned by removing the symbols and words written in non Arabic letters.
- For each corpus, the occurrence frequencies of all words are calculated and ordered in decreasing order. In this step, each corpus is analyzed separately because of size differences between the corpuses, and the fact that frequent words vary according to the nature of the corpus. Indeed, words such as (told us) "حدثنا / HdvnA", (he said) "قال /qAl " and (he prayed) " صلى / SlY ", which are very common in classical Arabic texts, are almost non-existent in the resolutions of the General Assembly of the United Nations.
- The next step consists to develop the first list of dictionary entries of the most frequent words. We have integrated the list of words related to each corpus and we have eliminated the diacritics that appear in some of them. Finally, a single word is stored in the case of words belonging to more than one list. The size of this dictionary has reached 16,200 words.
- Then, diacritics corresponding to dictionary entries are identified. For this, the entries are divided into two groups:

  - The first group consists of the words from the vowelized corpus (*Tashkeela*, NEMLAR and RDI). Every word of this group is accompanied by different lists of possible diacritics.
  - The second group consists of the remaining 1,200 words. Because this group contains many foreign words in Arabic language commonly used in contemporary texts and foreign proper names, these words are analyzed using the Aramorph Analyzer [19].This step is completed by performing a manual correction of the outputs of this analysis.

After making these steps, a dictionary of 16,200 most frequent words is obtained in the available corpuses. Each word of this dictionary is accompanied by a list of its different possible vowelizations. This lexicon is integrated in the process of morphological analysis program Alkhalil Morpho Sys. This has allowed immediately recognize possible vowelizations of the words belonging to this dictionary and this has led to faster processing.

### 3.1.3. Unanalyzed Words

During the morphological analysis step, some Arabic words are not analyzed due to incomplete coverage of the Alkhalil Analyzer databases. Indeed, for the authentic version of Alkhalil (version without the lexical basis of the most frequent words), the percentage of unanalyzed words is equal to 5.92%, while for the version including the dictionary of the most common words this percentage decreases to 2.74%.

## 3.2. Statistical Analysis

After the program has conducted a morphological analysis of text words, allowing to have the potential vowelizations for each word, the second step of the vowelization process is initiated. It consists of a statistical treatment based on the hidden Markov model and the Viterbi algorithm, and allows obtaining the most likely vowelization of words in the sentence. In what follows, the mechanisms of this model are recalled in detail. Let O={$o_1$,...,$o_M$} be a finite set of observations and let S = {$s_1$, ..., $s_N$} be a finite set of hidden states.

**Definition 1**

A first-order HMM is a double process $(X_t,Y_t)_{t\geq 1}$ where:

- $(X_t)_{t\geq 1}$ is a homogeneous Markov chain with values in the hidden states set S where:

$$\Pr(X_{t+1} = s_j / X_t = s_i, \ldots, X_1 = s_h) = \Pr(X_{t+1} = s_j / X_t = s_i) = a_{ij} \quad (1)$$

$a_{ij}$ is the transition probability from state $s_i$ to state $s_j$ and the matrix A = ($a_{ij}$) is called transition matrix.

- $(Y_t)_{t\geq 1}$ is an observable process taking values in the observation set O where:

$$\Pr(Y_t = o_k / X_t = s_i, Y_{t-1} = o_{k_{t-1}}, X_{t-1} = s_{i_{t-1}}, \ldots, Y_1 = o_{k_1}, X_1 = s_{i_1}) = \Pr(Y_t = o_k / X_t = s_i) = b_i(k) \quad (2)$$

$b_i(k)$ is the probability of observing $o_k$ given the state $s_i$ and the matrix B = ($b_i(t)$) is called transmission matrix.

### 3.2.1. First Model

In this model, the observations are unvowelized words of the sentence, and the hidden states are the vowelized words as in previous studies [1,10]. For example, the unvowelized word "دخل/dxl" is the observed state and the hidden state is one of the possible vowelized words such as (he came) " دَخَلَ / daxala " or (income) "دَخْلٌ/daxolN" given in the morphological analysis step (see Figure 3 below).

### 3.2.2. Second Model

The observations of this model are identical to those of the first model (unvowelized words), but the hidden states are the lists of possible diacritics in Arabic. So, they are vectors which have the same sizes as the observations (see Figure 4 below). To illustrate this definition, some examples of hidden states are given :

**Example 1:**

"َ, ْ, ِ, ُ": is the succession of short vowels (fatha), (Sukuun), (kasra) and (damma). It represents a potential hidden state of several observations related to words of four characters such as "تعرف / tErf", "نصبر / nSbr" and "يعرض / yErD". After applying these signs to these words, we get (she knows) "تَعْرِفُ / taEorifu", (we are patient) "نَصْبِرُ/ naSobiru" and (he displays) "يَعْرِضُ / yaEoriDu" .

## Example 2:

"ُ, َ, ِّ, َ, ٌ" : is the succession of five short vowels (damma), (fatha), (shadda) with (kasra), (fatha) and (tanween al-damma). To this list correspond several observations such as "معلمة / mElmp" and " مقدمة / mqdmp" and it allows us to have the following vowelized words (teacher) "مُعَلِّمَةٌ / muEal~imapN" and (introduction) "مُقَدِّمَةٌ / muqad~imapN".

## Example 3:

"َ, #, َ": is the succession of short vowel (fatha), the sign "#" and (fatha). The sign "#" represents the lack of diacritic sign on the corresponding character. This list allows to vowelize words such as "باع / bAE", "جاء / jA' " and "فاز / fAz" and so the following vowelized words are obtained : (he sold) "بَاعَ / baAEa", (he came) "جَاءَ /jaA'a" and (he won) "فَازَ / faAza".

It is obvious that the abstract nature of the hidden states of this model requires fewer parameters compared to first model which use the vowelized words as hidden states.

### 3.2.3. Viterbi Algorithm

In what follows, there is an explanation of the use of these models to vowelize Arabic texts. Suppose, for example, that there is an Arabic sentence $W=(w_1, w_2, ..., w_n)$ consisting of n words $w_i$. The series of observation is a set of unvowelized words $w_1, w_2, ..., w_n$. Let $C = \{c_1, c_2, ..., c_N\}$ be the set of the hidden states (vowelized words for the first model and possible lists of Arabic diacritic signs for the second model). Based on these assumptions, determining the correct vowelization at this level of analysis is to find a sequence of hidden states $(c_1^*, c_2^*, ..., c_n^*)$ in $C$ which satisfies:

$$(c_1^*,\ldots,c_n^*) = \underset{c_1\ldots c_n \in C}{\arg\max} \Pr(c_1\ldots c_n / w_1\ldots w_n). \quad (3)$$

Since

$$\Pr(c_1\ldots c_n / w_1\ldots w_n) = \frac{\Pr(w_1\ldots w_n / c_1\ldots c_n)\Pr(c_1\ldots c_n)}{\Pr(w_1\ldots w_n)} \quad (4)$$

The sequence $(c_1^*, c_2^*, ..., c_n^*)$ verifies:

$$(c_1^*,\ldots,c_n^*) = \arg\max \Pr(w_1\ldots w_n / c_1\ldots c_n)\Pr(c_1\ldots c_n) \quad (5)$$

The equation (5) can be written as follows:

$$(c_1^*,\ldots,c_n^*) = \underset{\substack{c_i^{j_i} \in C_i \\ 1\le i\le n}}{\arg\max} \Pr(w_1\ldots w_n / c_1^{j_1}\ldots c_n^{j_n})\Pr(c_1^{j_1}\ldots c_n^{j_n}) \quad (6)$$

Where $C_i = \{c_i^1,\ldots,c_i^{n_i}\}$ is for a first model (respectively second model) a set of the possible vowelized words (respectively the lists of diacritic signs of the possible vowelized words) obtained by morphological analysis of the words $w_i$. The equation (6) will be solved by seeking the most likely path in the network solutions obtained by morphological analysis of words out of context:

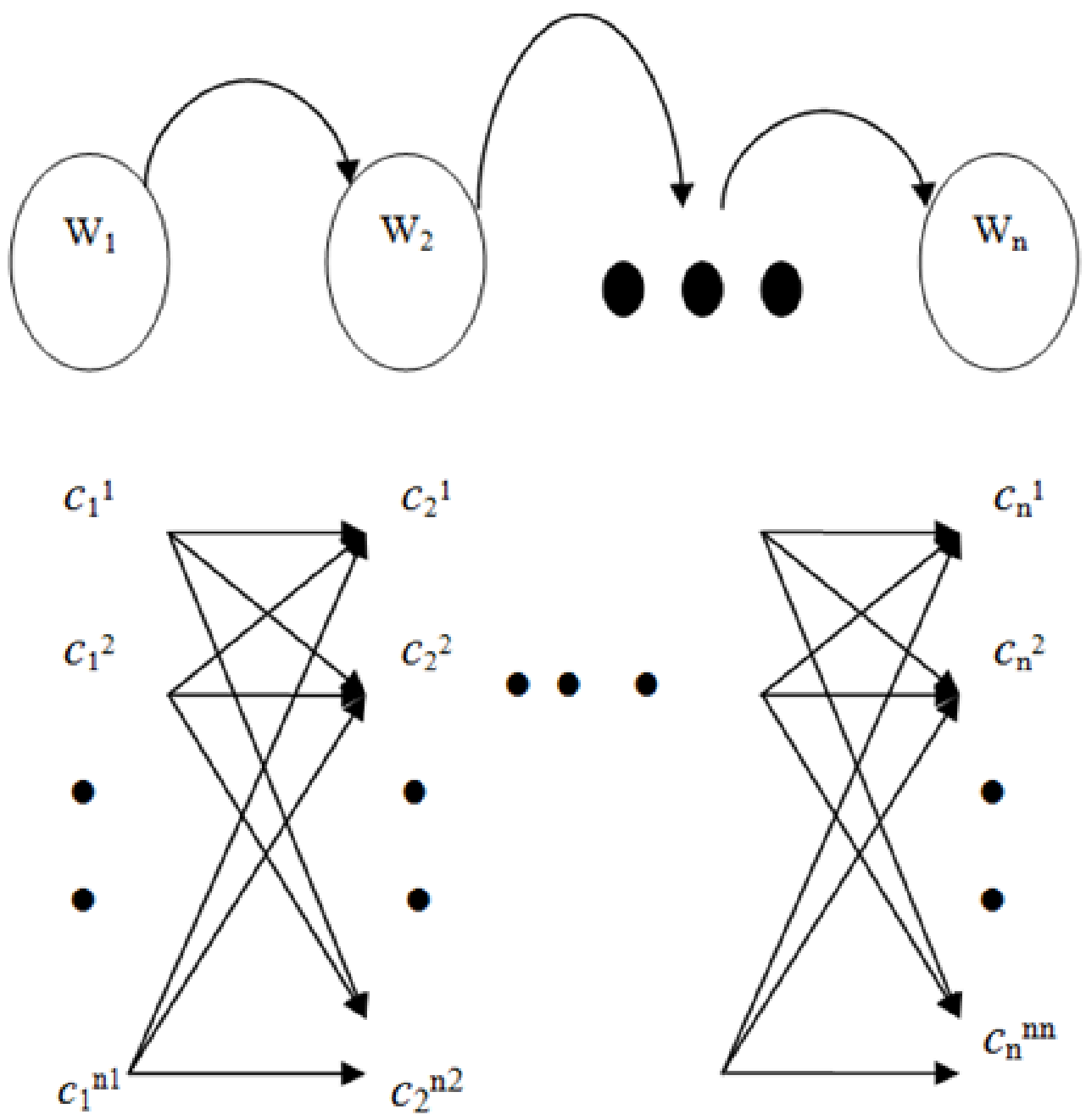

Figure 2. Network list of possible vowelizations obtained by morphological analysis of the sentence $W=(w_1, w_2, \dots, w_n)$

The Viterbi algorithm [23] identifies the optimal path in the network. It is based on the values of the function $\phi$ defined by:

$$\phi(t,c_t^k)=\max_{\substack{r_i^{j_i}\in\Re_i\\ 1\le i\le t-1}}\Big[\Pr(w_1,\dots,w_{t-1},w_t/c_1^{k_1},\dots,c_{t-1}^{k_{t-1}},c_t^k) \times\Pr(c_1^{k_1},\dots,c_{t-1}^{k_{t-1}},c_t^k)\Big] \quad (7)$$

Where $\phi(t,c_t^k)$ represents the probability of the partial path which passes through the list $c_t^k$ ($c_t^k$ belongs in the set of possible vowelizations of the word $w_k$). The equation (7) can be written as follows:

$$\begin{aligned}\phi(t,c_t^k) &= \max_{\substack{r_i^{j_i}\in\Re_i\\ 1\le i\le t-1}}\prod_{i=1}^{t-1}\Big[\Pr(w_i/c_i^{j_i})\times\Pr(c_i^{j_i}/c_{i-1}^{j_{i-1}})\Big]\times\Pr(w_t/c_t^k)\times\Pr(c_t^k/c_{t-1}^{j_{t-1}})\\ &=\Big(\max_{c_{t-1}^{j_{t-1}}\in C_{t-1}}\phi(t-1,c_{t-1}^{j_{t-1}})\times\Pr(c_t^k/c_{t-1}^{j_{t-1}})\Big)\Pr(w_t/c_t^k)\\ &=\Big(\max_{c_{t-1}^{j}\in C_{t-1}}\phi(t-1,c_{t-1}^{j})\times\Pr(c_t^k/c_{t-1}^{j})\Big)\Pr(w_t/c_t^k)\end{aligned} \quad (8)$$

The last formula (8) allows us to calculate the value of $\phi$ by induction. To identify the optimal path, the function $\psi$ is defined. It allows at any time t to store the vowelization which generates the largest value for the equation (8) above.

The function $\psi$ is defined by:

$$\psi(t,c_t^k)=\arg\max_{c_{t-1}^{j}\in C_{t-1}}\phi(t-1,c_{t-1}^{j})\Pr(c_t^k/c_{t-1}^{j}) \quad (9)$$

With the remark that $\psi(t, c_t^k) \in C_{t-1}$ . The relations (8) and (9) allow us to find the optimal path using the following decreasing Viterbi algorithm:

**Step 1 (initialization):**

For $1 \leq k \leq n_1$ , calculate $\psi(t, c_1^k)$ the probability that the sentence begins with $w_1$ accompanied by a hidden state $c_1^k$ .

**Step 2 (recursive computation):**

For $2 \leq t \leq n$ et $1 \leq k \leq n_t$,
calculate $\phi(t, c_t^k)$ and $\psi(t, c_t^k)$ from the following formulas:

$$\phi(t, c_t^k) = \left( \max_{c_{t-1}^j \in C_{t-1}} \phi(t-1, c_{t-1}^j) \times \Pr(c_t^k / c_{t-1}^j) \right) \Pr(w_t / c_t^k)$$

$$\psi(t, c_t^k) = \arg \max_{c_{t-1}^j \in C_{t-1}} \phi(t-1, c_{t-1}^j) \Pr(c_t^k / c_{t-1}^j)$$

**Step 3 (final state):**

$$\psi(n+1) = \arg \max_{c_n^j \in C_n} \phi(n, c_n^j)$$

**Step 4 (deduction of the optimal path):**

$$c_n^* = \psi(n+1)$$

For $t = n-1 : 1$

$$c_t^* = \psi(t, c_{t+1}^*)$$

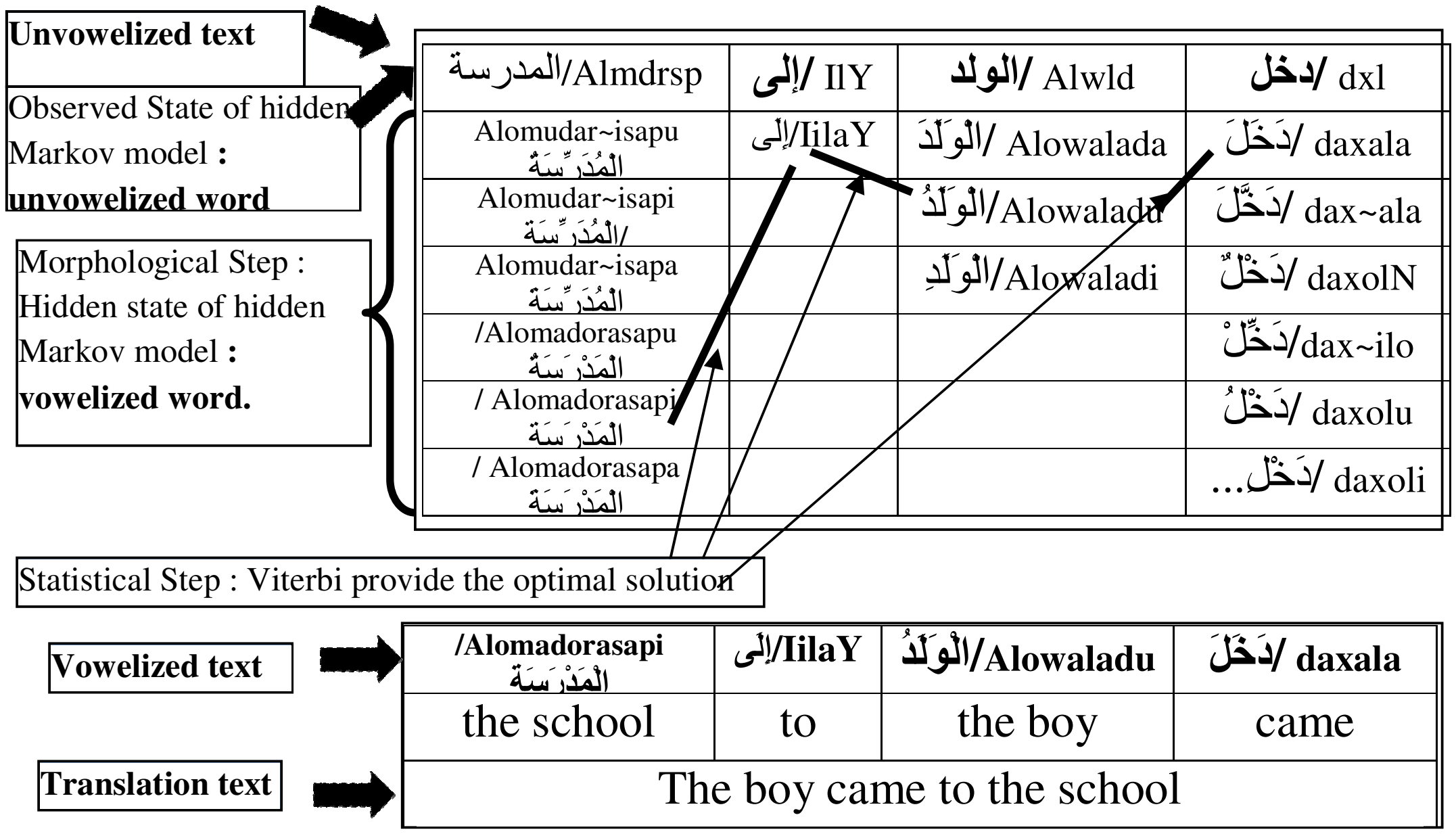

Figure 3. Using the Viterbi algorithm to find the optimal solution with the first model

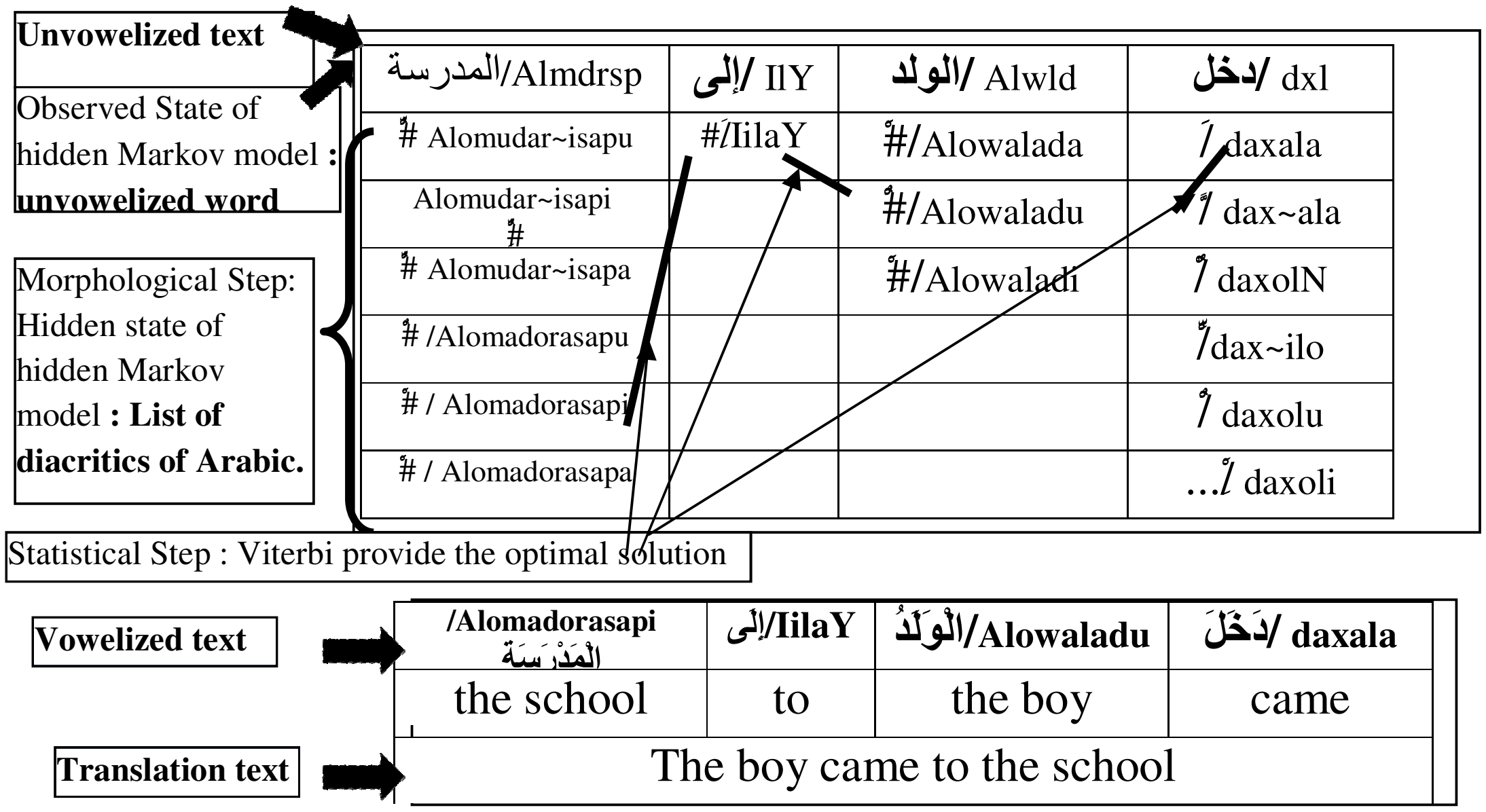

Figure 4. Using the Viterbi algorithm to find the optimal solution with the second model

## 4. TRAINING PHASE

The parameters of the statistical model defined above $A=(a_{ij})_{ij}$ and $B=(b_i(t))_{it}$ will be estimated during the training phase from representative linguistic corpora. Thus, if C $=\{Ph_1, \ldots, Ph_k\}$ is an Arabic corpus formed by M sentences $Ph_k$, then the training phase consists of estimating the parameters of the matrices A and B by the maximum likelihood method [15]. Indeed, if we put

- $n_i^k$ = the number of occurrences of the hidden state $c_i$ in the sentence $Ph_k$,
- $n_{ij}^k$ = the transition number from the state $c_i$ to the state $c_j$ in the sentence $Ph_k$,
- $m_{it}^k$ = the number of times that the unvowelized word $w_t$ correspond to the hidden state $c_i$ in the sentence $Ph_k$,

The coefficients $a_{ij}$ and $b_i(t)$ can be estimate by:

$$a_{ij} = \frac{\sum_{k=1}^{M} n_{ij}^k}{\sum_{k=1}^{M} n_i^k} \quad (10)$$

$$b_i(t) = \frac{\sum_{k=1}^{M} m_{it}^k}{\sum_{k=}^{M} n_i^k} \quad (11)$$

During the statistical phase, the unanalyzed words by Alkhalil are labelled by the label "unkown". Before proceeding to the Viterbi algorithm, an initial probability is assigned equal to $10^{-5}$ for all probabilities of the transition and transmission matrices relating to the state "unknown". This technical trick bypasses the problem of transition between unanalyzed words and the other words of the sentence. These unanalyzed words will not be vowelized by this statistical step of our system. So, for each unanalyzed word, another hidden Markov model is used which the observed states are the word letters and the hidden states are the diacritic signs.

The training phase was carried out with 90% of a corpus consisting of 2,463,351 vowelized words divided between NEMLAR corpus (460,000 words), (*Tashkeela*) corpus (780,000 words) and RDI corpus (1,223,351 words). The remaining 10% will be used in the test phase. The reason for non-use of million vowelized words available in the literature of classical Arabic is the desire to use a balanced corpus between contemporary texts and texts of classical Arabic. Thus, in parallel with contemporary texts of NEMLAR corpus and 13 books of RDI corpus, we selected 30 books of classical Arabic of (*Tashkeela*) corpus and 4 classic books from the RDI corpus, and then a random 10% of each book is picked. Finally, operations are used which consist in segmenting texts into sentences, classifying the words of these sentences and deducing the corresponding vowelized words necessary to estimate model parameters.

For unanalyzed words in morphological step, a model based on the characters is used. Each word will be fragmented into characters and diacritics. Transitions are between the letters without diacritics and emissions are calculated from the observed states (characters) and the hidden states (diacritics). The hidden Markov model was generated by learning 90% on the same corpus of 2.46 million words.

## 5.TEST PHASE

The characteristics of the used equipment are a PC with 2 CPU Intel (R) Pentium (R) 1.9 GHz and RAM with 3Gigats bytes. The used environment is the operating system Windows 7. The application has been extended to 1,024 Megabytes memory. To evaluate the performance of our system, the error rate at the word level WER (WER: Word Error Rate) and the error rate at the character level DER (DER: Diacritic Error Rate) are calculated. For each of these two types of errors, we compute both the error rate which takes into account the diacritic sign of the last character and the one that ignores this sign. Thus, the 4 following error rates are calculated:

- WER1: calculates the rate of words which are vowelized wrongly by the program taking into account the diacritic sign of the last character.
- WER2: calculates the rate of words which are vowelized wrongly by the program while ignoring the diacritic sign of the last character.
- DER1: calculates the rate of characters which are vowelized wrongly by the program taking into account the diacritic sign of the last character.
- DER2: calculates the rate of characters which are vowelized wrongly by the program while ignoring the diacritic sign of the last character.

The test was focused on the test corpus composed of 199,197 words and completely independent of the learning corpus.

It is recalled that the original version of the morphological Analyzer Alkhalil Morpho Sys is modified by including the dictionary of the most frequent words. To evaluate the contribution of this change, our vowelization system is tested at the first using the original version of Alkhalil Morpho Sys in the morphological analysis phase, and thereafter one that uses the modified version of Alkhalil Morpho Sys.

### 5.1. Evaluation without the Dictionary of the Most Frequent Words

The Table 1 below shows the evaluation results of our vowelization system which use in the first phase (morphological analysis phase) the original version of the Analyzer Alkhalil Morpho Sys. The second line of Table 1 indicates the execution time and the four error rates of the first model in which the hidden states are vowelized words. The third line of Table 1 is devoted to the execution time and the four error rates of the second model in which the hidden states are the lists of possible diacritic signs of Arabic.

Table 1. Evaluation of the two models of the test corpus when AlKhalil Analyzer is used without the dictionary of the most frequent words

| Model | Execution time Number of words/s | WER1(%) | WER2(%) | DER1(%) | DER2(%) |
|---|---|---|---|---|---|
| The first model | 90.34 | 30.88 | 16.91 | 11.38 | 6.01 |
| The second model | 86.85 | 28.31 | 16.69 | 10.56 | 5.96 |

These results show that the system performance obtained with the second model are better than those relating to the first model. The first system is faster than the second (90.34 words are vowelized per second for the first model against 86.56 words per second for the second). This can be explained by the fact that the transmission matrix B is probabilistic (all values of B are equal to 1 or 0).

However, the error rate at the word level (WER1) is around 30.88% for the first model and 28.31% for the second model. This error rate decreases by almost half if the diacritic of the last character is ignored (WER2). The error rate DER1 relating to all characters of the text is about 11% and decreases, as in the case of words, by almost half in the case where the diacritic sign of the last character is omitted.

## 5.2 Evaluation with the Dictionary of the Most Frequent Words

Table 2 below shows the evaluation results of our vowelization system using in the first phase the modified version of Alkhalil Morpho Sys which include the dictionary of the most frequent words. As in Table 1, the second line of Table 2 shows the execution time and the four error rates of the first model in which the hidden states are vowelized words. The third line of Table 2 is devoted to the execution time and the four error rates of the second model in which the hidden states are the lists of possible diacritic signs of Arabic.

Table 2. Evaluation of the two models of the test corpus when AlKhalil Analyzer is used with the dictionary of the most frequent words

| Model | Execution time Number of words/s | WER1(%) | WER2(%) | DER1(%) | DER2(%) |
|---|---|---|---|---|---|
| The first model | 109.63 | 21.11 | 9.93 | 7.37 | 3.75 |
| The second model | 105.34 | 21.41 | 10.59 | 7.47 | 3.95 |

Unlike the previous paragraph, the results obtained with the first model are better than those relating to the second model. The error rate at the word level (WER1) is around 21% (21.10% for the first model and 21.41% for the second model), and it decreases by more than half if the diacritic of the last character is ignored (WER2).

The error rate DER1 relating to all characters of the text is about 7% and decreases, as in the case of words, by almost half in the case where the diacritic sign of the last character is omitted.
The use of the dictionary of the most frequent words makes on the one hand the system faster (the first model of this system vowelized 109.63 words per second against 90.34 words per second for the system without the dictionary). On the other hand, it has allowed to significantly improving system performance. Indeed, the error rates WER1, WER2, DER1 and DER2 relating the use the original version of Alkhalil Morpho Sys in the system, and which are respectively equal to 30.88%, 16.91%, 11.38% and 6.01%, decreased to 21,11%, 9.93%, 7.37% and 3.75 respectively, after having inserted into Alkhalil Morpho Sys the dictionary of the most frequent words.

## 5.3. Comparison of results

To compare our results with those of other systems in the literature, Table 3 shows the different error rates. However, as these systems have not been tested on the same corpus, it should take the conclusions with some caution.

Table 3. Comparing the performance of some systems with our vowelizer

| Vowelization system | WER1(%) | WER2(%) | DER1(%) | DER2(%) |
|---|---|---|---|---|
| Nelken et al. (2005) [10] | 23.61 | 7.33 | 12.79 | 6.35 |
| (Al Ghamdi et al. 2010) [6] | 46.83 | 26.03 | 13.83 | 9.25 |
| Bebah et al (2012) [12] | / | 20.5 | / | 8.2 |
| Our vowelizer (first model) | 21.11 | 9.93 | 7.37 | 3.75 |

## 5.4. Synthesis

It emerges from these evaluations the following conclusions:

- The integration of the dictionary of the most frequent words in Alkhalil Analyzer has allowed to make more efficient the vowelization systems. Indeed, the error rate WER1 decreased by 31% for model 1 (it decreased from 30.88 to 21.11) and 25% for model 2 (it decreased from 28.31 to 21.41). Similarly, the error rate WER2 was decreased by 41% for model 1 (it decreased from 16.91 to 9.92) and 42% for model 2 (it decreased from 10.59 to 6.1).
- It also makes the system faster. Indeed, the system using the original version of Alkhalil Morpho Sys vowelized 90.34 words per second for the first model and 86.65 words per second for the second model, when that using the dictionary vowelized 109.63 words per second for the first model and 105.34 words per second for the second model.
- When Alkhalil analyzer is used without the dictionary of the most frequent words, the error rates obtained with model 1 are slightly less good than those of model 2 (see Table 1). The trend is reversed after integrating this dictionary in Alkhalil Analyzer (see Table 2).
- The first vowelization model is faster than the second independently of integration or not of the dictionary of the most frequent words in the morphological step (without the dictionary, the first model analysis 109 words per second against 90 for the second, and with the dictionary, the first model analysis 105 words per second against 86 for the second). This goes against our expectations. Indeed, the sizes of the transition and transmission matrices of the second model (size(A) = 21,267 x 21,267 and size(B) = 21,267 x 145,469) are much smaller than that of the first model (size(A) = 224,414 x 224,414 and size(B) = 224,414 x 145,469). So it was natural to expect that the second model will be the fastest. This unexpected finding may be explained by the probabilistic nature of the transmission matrix B of the first model (the coefficients of the matrix B are all equal to 1 or 0).
- The error rate WER1decreases by almost half if the diacritic of the last character is ignored (error rate WER2). So, almost half of the errors are syntactic errors.
- It is recalled that for computing the error rates WER1 and WER2, a word is considered correctly vowelized by our system if there is total agreement between the vowels given by the vowelization system and those in the corpus. As the corpus is marred by several errors (including spelling errors), a negative impact is generated both in the learning phase (to estimate the matrices A and B) and in the test phase. This partly explains the obtained values for WER1 and WER2.

- To analyze the performance of each of the two steps of our system (the morphological analysis step and the statistical step), the error share for each of these two steps will be identified. It should be noted that the vowelization error of a word is due to the one of the following three reasons:
  – Alkhalil Analyzer does not analyze the word.
  – Alkhalil Analyzer analyzes the word but the right solution (the solution in the context) is not included in the analysis outputs.
  – Viterbi algorithm does not identify the right solution even if it is one of the outputs provided by Alkhalil Analyzer.

Table 4 shows the distribution of the error rate at the word level WER1 according to each of these reasons when the original version of Alkhalil Morpho Sys is used in the morphological step.

Table 4. Distribution of the error rate WER1 on the test corpus relating to the system which uses the original version of AlKhalil Analyzer

| **From words not correctly vowelized, (%) of those** | **First model** | **Second model** |
|---|---|---|
| unanalyzed by AlKhalil | 15.06 | 16.42 |
| for which the right solution does not belong to the analysis outputs | 25.34 | 27.64 |
| that the right solution belongs to the analysis outputs by not given by Viterbi algorithm | 59.6 | 55.94 |

It is clear from these statistics that between 40% and 45% vowelization errors (depending on model) are due to the analysis morphological step and the rest is a consequence of the statistical step. Distributions for the other kinds of errors (WER2, DER1 and DER2) look like those shown in Table 4.

## 6. CONCLUSION AND OUTLOOK

Automatic vowelization systems based on hybrid approaches that combine morphological analysis and hidden Markov models are presented in this article. These approaches differ from other hybrid approaches to linguistic and statistical levels. At the linguistic level, Alkhalil Morpho Sys which is an open source morphological Analyzer is used. Our system uses the outputs of this Analyzer and extracts possible vowelizations out of context for each word.
The integration of this Analyzer in our vowelization system required the addition of a lexical database containing the most frequent words in Arabic language. This has allowed to adjust their diacritic signs and to circumvent problems of slow due to the high number of solutions proposed by Alkhalil Analyzer. This lexicon has been generated from a database of more than 250 million words from eight Arab corpuses available on the Internet.

Statistically, two Markov models are used whose observations are unvowelized Arabic words and the hidden states are for the first model the vowelized words and for the second the lists of diacritic marks of the words. Similarly, in the training phase, a varied corpus was used containing classical and contemporary Arabic texts and whose size exceeds 2.46 million words. Finally, the evaluation results obtained are very encouraging, especially for the system incorporating the lexical database. Our system can be improved by acting on the following complementary levels:

- Improve the performance of Alkhalil system so that these outputs will be more accurate and its coverage broader (reduce the error rate of the unanalyzed words by the system).
- Enrich the corpus used in the learning and testing phases and correct any errors in the corpus.
- Exploiting syntactic information provided by the Alkhalil Analyzer to reduce the error rate relating to the vocalization of the last word character (it is recalled that almost half of the errors are at the last character of the word).
- As the corpus used in the learning phase cannot cover all the Arabic language, some coefficients of the estimated transition and transmission matrices are equal to zero. In order to improve the performance of our systems, some smoothing techniques studied in [24] will be used to estimate these coefficients.

**Authors**

Mohamed Ould Abdallahi Ould Bebah

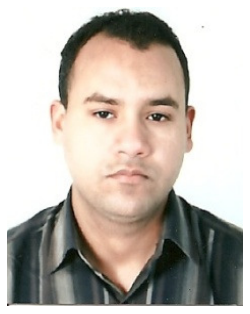

Researcher at Doha Institute for Graduate Studies since 2013. "Doctorat" in Computer Sciences, Mohamed I University, Oujda, Morocco, 2013. "DESA" in "Numerical Analysis, Computer science and Signal Processing" from Mohamed I University, 2005. Member of Arabic NLP unit, LaRI Laboratory, Mohamed I University since 2005. Member of Language Studies unit at the Center of Social and Human Studies and Researches (CERHSO) in Oujda since 2005. Member of Arabic Language Engineering Society in Morocco (ALESM) since 2012.

Amine CHENNOUFI

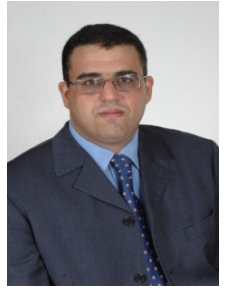

Master in Computer Sciences from Mohamed I University, Oujda, Morocco 2010. Engineering degree in Meteorology from the National School of Meteorology ENM in Toulouse in France since 1994. Since January 2011, He prepares his PhD thesis in Arabic Natural Language Processing within the Computer Research Laboratory (LaRI). His research interests are especially in Automatic vowelization of Arabic language. Professionally he is the responsible of Meteorological Centre of Oujda Airport.

Azzeddine Mazroui

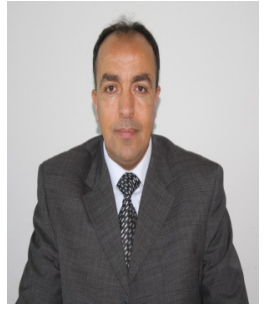

"Doctorat d'Etat" in Numerical Analysis, University Mohammed I Morocco, 2000. PHD in Probability and Statistics, Pierre & Marie Curie University France, 1993. Professor of mathematics and Computer Sciences in University Mohammed I. Member of Computer Research Laboratory (LaRI). Director of the ANLP unit in the LaRI laboratory.

Abdelhak Lakhouaja

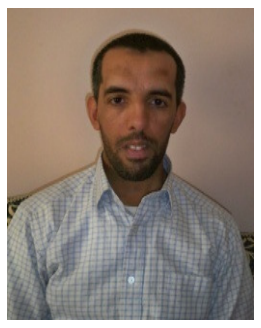

"Doctorat d'Etat" in Computer Sciences, University Mohammed I Morocco, 2000. Professor of Computer Sciences in University Mohammed I. Member of Computer Research Laboratory (LaRI). Cofounder of the ANLP unit in the LaRI laboratory.